\relax
\documentclass[letterpaper]{article} %DO NOT CHANGE THIS
\usepackage{aaai18}  %Required
\usepackage{times}  %Required
\usepackage{helvet}  %Required
\usepackage{courier}  %Required
\usepackage{url}  %Required
\usepackage{graphicx}  %Required
\newcommand{\model}{\mbox{TPRN}\,}
\newcommand{\modell}{\mbox{TPRN}}
\newcommand{\R}{\mathbb{R}}
\newcommand{\Q}{\mc{Q}}

\newcommand{\aft}{\mb{a}_\mr{S}\!^{(t)}}
\newcommand{\art}{\mb{a}_\mr{R}\!^{(t)}}
\newcommand{\artT}{\mb{a}_\mr{R}\!^{(t)\top}}

\newcommand{\mb}[1]{\mathbf{#1}}
\newcommand{\mr}[1]{\mathrm{#1}}

\newcommand{\mc}[1]{\mathcal{#1}}
\usepackage{amsfonts}
\usepackage{amsmath}
\usepackage{booktabs}
\usepackage{color}
\usepackage{subfigure}
\graphicspath{{./Figures/}}
\begin{document}
\title{Question-Answering with Grammatically-Interpretable Representations}
%%Commented the author info for blind review purposes.
\author{Hamid Palangi, Paul Smolensky, Xiaodong He, Li Deng\\
\texttt{\{hpalangi,psmo,xiaohe\}@microsoft.com}, \texttt{l.deng@ieee.org}\\
Microsoft Research AI\\
Redmond, WA \thanks{This work was carried out while PS was on leave from Johns Hopkins University. LD is currently at Citadel.}\\
}
\maketitle
\begin{abstract}
We introduce an architecture, the Tensor Product Recurrent Network (TPRN).
In our application of TPRN, internal represent\-ations---learned by end-to-end optimization in a deep neural network performing a textual question-answering (QA) task---can be interpreted using basic concepts from linguistic theory. No performance penalty need be paid for this increased interpretability: the proposed model performs comparably to a state-of-the-art system on the SQuAD QA task.
The internal representation which is interpreted is a Tensor Product Representation: for each input word, the model selects a symbol to encode the word, and a role in which to place the symbol, and binds the two together. The selection is via soft attention. The overall interpretation is built from interpretations of the symbols, as recruited by the trained model, and interpretations of the roles as used by the model. We find support for our initial hypothesis that symbols can be interpreted as lexical-semantic word meanings, while roles can be interpreted as approximations of grammatical roles (or categories) such as subject, wh-word, determiner, etc. Fine-grained analysis reveals specific correspondences between the learned roles and parts of speech as assigned by a standard tagger \cite{StanfParser}, and finds several discrepancies in the model's favor. In this sense, the model learns significant aspects of grammar, after having been exposed solely to linguistically unannotated text, questions, and answers: no prior linguistic knowledge is given to the model. What is given is the means to build representations using symbols and roles, with an inductive bias favoring use of these in an approximately discrete manner. 
\end{abstract}

\section{Introduction: Minding the gap}
\label{problem}

The difficulty of explaining the operation of deep neural networks begins with the difficulty of 
interpreting the internal representations learned by these networks.
This problem fundamentally derives from the incommensurability between, on the one hand, the continuous, numerical representations and operations of these networks and, on the other, meaningful interpretations---which are communicable in natural language through relatively discrete, non-numerical conceptual categories structured by conceptual relations. 
This gap could in principle be reduced if deep neural networks were to incorporate internal representations that are directly interpretable as discrete structures; the categories and relations of these representations might then be understandable conceptually. 

In the work reported here, we describe how approximately discrete, structured distributed representations can be embedded within deep networks, their categories and structuring relations being learned end-to-end through performance of a task. 
Applying this approach to a challenging natural-language question-answering task, we show how the learned representations can be understood as approximating syntactic and semantic categories and relations. 
In this sense, the model we present learns significant aspects of syntax/semantics, recognizable using the concepts of linguistic theory, after having been exposed solely to linguistically unannotated text, questions, and answers: no prior linguistic knowledge is given to the model. 
What \textit{is} built into the model is a general capacity for distributed representation of structures, 
%and a soft inductive bias favoring discreteness in the deployment of this capacity.
and an inductive bias favoring discreteness in its deployment.

Specifically, the task we address is question answering for the SQuAD dataset \cite{squad}, in which a text passage and a question are presented as input, and the model's output identifies a stretch within the passage that contains the answer to the question. 
In our view, SQuAD provides a sufficiently demanding QA task that showing interpretability of our proposed type of distributed structural representation in a QA system that successfully addresses SQuAD provides meaningful evidence of the potential of such representations to enhance interpretability of large-scale QA systems more generally.

The proposed capacity for distributed representation of structure is provided by Tensor Product Representations, TPRs, in which a discrete symbol structure is encoded as a vector systematically built---through vector addition and the tensor product---from vectors encoding symbols and vectors encoding the roles each symbol plays in the structure as a whole \cite{TPAI,THM,SGM}.
The new model proposed here is built from the {\sc BiDAF} model proposed in \cite{BIDAF} for question answering. 
We replace a bidirectional RNN built from LSTM units 
\cite{hochreiter1997long}
with one built from TPR units; the architecture is called the \emph{Tensor Product Recurrent Network}, \modell.
\model learns the vector embeddings of the symbols and roles, and learns which abstract symbols to deploy in which abstract roles to represent each of the words in the text-passage and query inputs. 

We show how the structural roles that \model  learns can be interpreted through linguistic concepts at multiple levels: morphosyntactic word features, parts of speech, phrase types, and grammatical roles of phrases such as subject and object. 
The match between standard linguistic concepts and \model's internal representations is approximate.
%and we identify several discrepancies in the model's favor.

The work reported here illustrates how learning to perform a typical natural language task can lead a deep learning system to create representations that are interpretable as encoding abstract grammatical concepts without ever being exposed to data labelled with anything like grammatical structure. 
It is commonly accepted among language acquisition researchers that it is in this type of setting that children typically learn their first language, so the work lends plausibility to the hypothesis that abstract notions of linguistic theory do  describe representations in speakers' minds---representations that are learned in the service of performing tasks such as question-answering which (unlike, say, a parsing task) do not explicitly necessitate any such structure.

The remainder of the paper is structured as follows.
Section \ref{model} provides some background while Section \ref{TPR} introduces TPR and details how it is used in the general \model architecture we propose here.
Experimental results applying TPRN to SQuAD are presented in Section \ref{experiments}.
The heart of the paper is Section \ref{interpretation} which addresses interpretation of the representations learned by TPRN.
Section \ref{related_work} discusses related work and Section \ref{conclusion} concludes.

\section{The Model}
\label{model}
The proposed \model architecture is built in TensorFlow \cite{tensorflow2015-whitepaper} on the {\sc BiDAF} model proposed in \cite{BIDAF}.
{\sc BiDAF} is constructed from 6 layers: 
a character embedding layer using CNNs, 
a word embedding layer using GloVe vectors \cite{GLOVE}, 
a phrase embedding layer using bidirectional LSTMs for sentence embedding \cite{hpTASLP}, 
an attention flow layer using a special attention mechanism, 
a modeling layer using LSTMs, 
and an output layer that generates pointers to the start and end of an answer in the paragraph.
(See Fig. 1 of \cite{BIDAF}.) 

\begin{figure*}[ht!]
\begin{center}
\includegraphics[width=\textwidth]{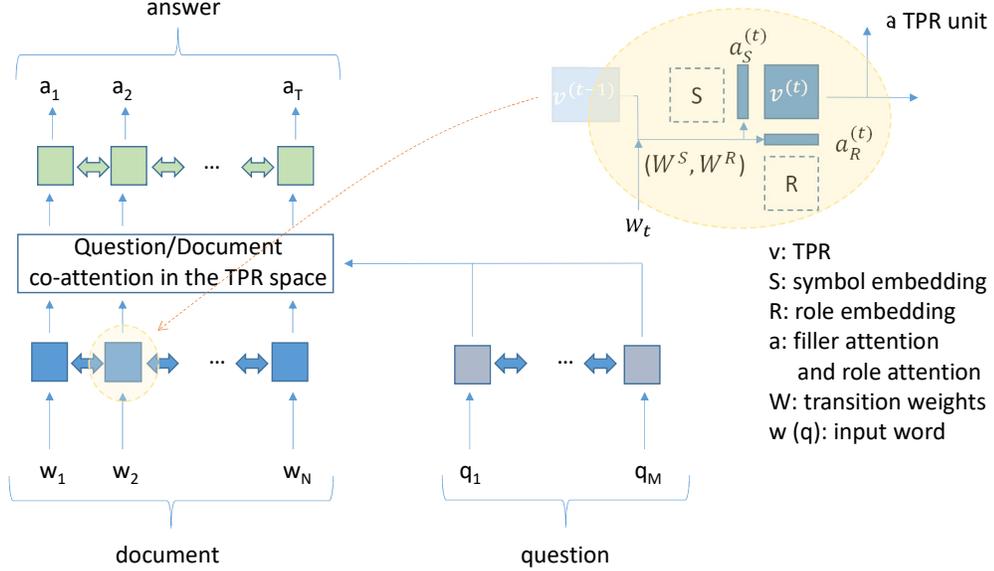}
\vspace{-3cm}
\caption{Block diagram of the proposed model.}
%\vspace{-.5cm}
\label{fig:TPR}
\end{center}
\end{figure*}

 \model replaces the LSTM cells forming the bidirectional RNN in the phrase embedding layer with recurrent TPR cells, described next: 
%The block diagram of the proposed model is presented in Fig. \ref{fig:TPR}.
see Fig. \ref{fig:TPR}.

\section{TPRN: The Tensor Product Recurrent Network}
\label{TPR}

%Although the approach generalizes in obvious ways, for conciseness, here we describe the particular model whose performance is described in section \ref{experiments} and whose representations are interpreted in section \ref{interpretation}. 
This \model model enables the phrase-embedding layer of the model to decide, for each word, how to encode that word by selecting among $nSymbols$ symbols, each of which it can choose to deploy in any of $nRoles$ slots in an abstract structure. 
The symbols and slots have no meaning prior to training.
We hypothesized that the symbol selected by the trained model for encoding a given input word will be interpretable in terms of the lexical-semantic content of the word (e.g., \emph{Australia} refers to a place) while the slots will be interpretable as grammatical roles such as subject/agent, object/patient, question-restrictor phrase. 
%(e.g., in \emph{J loves K}, \emph{J} plays a subject/agent role, \emph{K} plays the object/patient role, and \emph{loves} plays the verb/predicate role). 
In Section \ref{interpretation}, we will test this hypothesis; we will henceforth refer to ``roles'' rather than ``slots''.
In other words, our hypothesis was that the particular word tokens for which a given symbol was selected would form a lexical-semantically-related class, and the particular word tokens for which a given role was selected would form a grammatically-related class.

To function within the network, the symbols and roles must each be embedded as vectors; assume that we use vectors of dimension $dSymbols$ and $dRoles$ for symbols and roles respectively. These embedding vectors are designed by the network, i.e., they are learned during training. The network's parameters, including these embeddings, are driven by back-propagation to minimize an objective function relevant to the model's question-answering task. The objective function includes a standard cross-entropy error measure, but also \emph{quantization}, a kind of regularization function biasing the model towards parameters which yield decisions that select, for each word, a single symbol in a single role: the selection of symbols and roles is soft-selection, and we will say that the model's encoding assigns, to the $t^{th}$ word $w^{(t)}$, a \emph{symbol-attention vector} $\aft$ and a \emph{role-attention vector} $\art$.

The quantization term in the objective function pushes towards attention vectors that are 1-hot.
We do not impose this as a hard constraint because our fundamental hypothesis is that by developing \emph{approximately} discrete representations, the model can benefit from the advantages of discrete combinatorial representations for natural language, without suffering their disadvantage of rigidity. 
We note that while the attention vectors are approximately 1-hot, the actual representations deployed are attention-weighted sums of fully distributed vectors arising from distributed encodings of the symbols and distributed embeddings of the roles.

In the encoding for $w^{(t)}$, the vector $\mb{s}^{(t)}$ encoding the symbol is the attention-weighted sum of the $nSymbols$ possible symbols:  
$\mb{s}^{(t)} = \sum_{j=1}^{nSymbols} [\aft]_{j} \mb{s}_{j} = \mb{S} \aft$ 
where $\mb{s}_{j}$ is the embedding of the $j^{th}$ symbol in $\R^{dSymbols}$, which is the $j^{th}$ column of the \emph{symbol matrix} $\mb{S}$.
Similarly, the vector encoding the role assigned to $w^{(t)}$ is $\mb{r}^{(t)} = \sum_{k=1}^{nRoles} [\art]_{k} \mb{r}_{k} = \mb{R} \art$,
with $\mb{r}_{k}$  the embedding of the $k^{th}$ symbol in $\R^{dRoles}$ and the $k^{th}$ column of the \emph{role matrix} $\mb{R}$.
%(The dimensions of symbol- and role-vectors need not be the same.) 
Since the symbol $\{ \mb{s}_{j} \}_{j=1:nSymbols}$ and role $\{ \mb{r}_{k} \}_{k=1:nRoles}$ vectors are unconstrained, they generally emerge from the learning process as highly distributed; that is even more true of the overall representations $\{ \mb{v}^{(t)} \}$, as we now see.

The activation vector $\mb{v}^{(t)}$ that encodes a single word $w^{(t)}$ combines the word's symbol-embedding vector, $\mb{s}^{(t)}$, and its role-embedding vector, $\mb{r}^{(t)}$, via the outer or tensor product: $\mb{v}^{(t)} = \aft \artT = \aft \otimes \art $. We say that $\mb{v}^{(t)}$ is the \emph{tensor product representation (TPR) of the binding of symbol $\mb{s}$ to the role $\mb{r}$}.
A convenient expression for  $\mb{v}^{(t)}$ is:
\begin{align}
\label{eq:TPRforward}
\mb{v}^{(t)} &\equiv \mb{s}^{(t)} (\mb{r}^{(t)})^{\top}
= \left( \mb{S} \aft \right) \left( \mb{R} \art \right)^{\top}
\\ \nonumber
&= \mb{S}  \left( \aft \artT \right) \mb{R}^{\top} 
= \mb{S} \mb{B}^{(t)} \mb{R}^{\top} 
\end{align}
The matrix $\mb{B}^{(t)} \equiv \aft \artT$ is the \emph{binding matrix} for word $w^{(t)}$, which encodes the (soft) selection of symbol and role for $w^{(t)}$.
This matrix has dimension $nSymbols \times nRoles$; the actual representation sent to deeper layers, $\mb{v}^{(t)}$, has 
$dSymbols \times dRoles$ embedding dimensions. 
(In the particular model discussed below, the dimensions of $\mb{B}$ and $\mb{v}^{(t)}$ 
are respectively $100 \times 20$ and $10 \times 10$. )

$\aft$ and $\art$ in (\ref{eq:TPRforward}) are defined as:
\begin{align}
\label{eq:aF}
\aft &= f(\mathbf{W}^S_{in}\mathbf{w}^{(t)} + \mathbf{W}^S_{rec}vec(\mathbf{v}^{(t-1)}) + \mathbf{b}^S) \\ 
\label{eq:aR}
\art &= f(\mathbf{W}^R_{in}\mathbf{w}^{(t)} + \mathbf{W}^R_{rec}vec(\mathbf{v}^{(t-1)}) + \mathbf{b}^R) 
\end{align}
where $vec(.)$ is the vectorization operation, $f(.)$ is the logistic sigmoid function, $\mathbf{w}^{(t)}$ is the $t^{th}$ word and $\mathbf{b}$ is a bias vector. 
Equation \eqref{eq:TPRforward} is depicted graphically in the `TPR unit' insert in Fig. \ref{fig:TPR}.
During $I \rightarrow O$ computation, in the forward-directed RNN, the representation $\mb{v^{(t-1)}}$ of the previous word is used to compute the attention vectors $\aft$, $\art$ which in turn are used to compute the representation  $\mb{v^{(t)}}$ of the current word. 
(The same equations, with the same transition weights and biases, apply to the words in both the passage and the query.)

Because each word is represented (approximately) as the TPR of a single symbol/role binding, we can interpret the internal representations of \model's phrase-embedding layer once we can interpret the symbols and roles it has invented. 
Such interpretation is carried out in the Section \ref{interpretation}.

The interest in TPR lies not only in its interpretability, but also in its power.
The present \model model incorporates TPR to only a modest degree, but it is a proof-of-concept system that paves the way for future models that can import the power of general symbol-structure processing, proven to be within the scope of full-blown TPR architectures \cite{THM,PhilTr}. 
\model is designed to scale up to such architectures; design decisions such as factoring the encoding as  $\mb{v}^{(t)} = \aft \artT = \aft \otimes \art$  are far from arbitrary:
they derive directly from the general TPR architecture.

As the name \model suggests, the novel representational capacity built into \model is an RNN built of TPR units: a forward- and a backward-directed RNN in each of which the word $w^{(t)}$ generates an encoding which is a TPR: $\mb{v}^{(t)} = \mb{S} \mb{B}^{(t)} \mb{R}^{\top}$; 
the binding matrix $\mb{B}^{(t)}$ varies across words, but a single symbol matrix $\bf{S}$ and single role matrix $\bf{R}$ apply for all words $\{ w^{(t)} \}$. Both the $\bf{S}$ and $\bf{R}$ matrices are learned during training.

It remains only to specify the quantization function $\Q$ (\ref{Q}) which is added (with weight $c_Q$) to the cross-entropy to form the training objective for \model:
$\Q$ generates a bias favoring attention vectors $\aft$ and $\art$ that are 1-hot.
\begin{align}
\label{Q}
&\Q = \Q_a(\aft) + \Q_a(\art)\\ \nonumber
&\Q_a(\mb{a}) =   \Sigma_i (a_i)^2 (1 - a_i)^2 + \left(  \Sigma_i (a_i)^2 - 1 \right)^2
\end{align}

The first term of $\Q_{a}$ is minimized when each component of $\mb{a}$ satisfies $a_{i} \equiv [\mb{a}]_i \in \{0, 1\}$; the second term is minimized when $\| \mb{a} \|_2^{2} = 1$. 
The sum of these terms is minimized when $\mb{a}$ is 1-hot
\cite{cho_bifurcation_2016}.
%[P. Tupper, personal communication; \citet{ChoSmo16}]
$\mc{Q}$ drives learning to produce weights in the final network that generate $\aft$ and $\art$ vectors that are approximately 1-hot, but there is no mechanism within the network for enforcing (even approximately) 1-hot vectors at $ I \rightarrow O$ computation (inference) time.

\section{Experiments}
\label{experiments}

In this section, we describe details of the experiments applying the proposed \model model to the question-answering task of the Stanford's SQuAD dataset \cite{squad}. 
The results of primary interest are the interpretations of the learned representations, discussed at length in the next section.
%\footnote{After the double-blind AAAI reviewing process, we will add a link to a GitHub repo of our source code here.} 

The  goal of this work is not to beat the state-of-the-art system on SQuAD (at the time of writing this paper, DCN+ from Salesforce Research), but to create a high-performing question-answering system that is interpretable, by exploiting TPRs. 
%Therefore, as long as we have a system with TPR cells that performs reasonably well, we can use it to provide proof of concept for the interpretability claims in this work. 
%Further exploration of hyper-parameters and extended TPR architectures promise to outperform the state-of-the-art system on SQuAD; this is ongoing work and outside the scope of this paper. 

SQuAD is a reading comprehension dataset for question answering. It consists of more than 500 Wikipedia articles and more than 100,000 question-answer pairs about them, which is significantly larger than previous reading comprehension datasets \cite{squad}. The questions and answers are human-generated. 
The answer to each question is determined by two pointers into the passage, one pointing to the start of the answer and the other one pointing to its end. 
Two metrics that are used to evaluate models on this dataset are Exact Match (EM) and F1 score.

For the experiments, we used the same settings reported in \cite{BIDAF} for all layers of \model except the phrase embedding layer, which 
%(Note that these hyperparameters had previously been optimized for the original model, and were not re-optimized for \modell.)
is replaced by our proposed recurrent \model cells. The full setting of the \model model for experiments is as follows:
\begin{itemize}
\item Questions and paragraphs were tokenized by the PTB tokenizer.
\item The concatenation of word embedding using GloVe \cite{GLOVE} and character embedding using Convolutional Neural Networks (CNNs) was used to represent each word. 
The embedding vector for each character was of length 8 (1-D input to the CNN) and the output of the CNN's max-pooling layer over each word was a vector of length 100. 
The embedding size of word embedding using GloVe was also set to 100. 
\item For the interpretability experiments reported in Section \ref{interpretation}, the hyperparameter values used for the TPRN cell were $nSymbols = 100$ symbols and $nRoles = 20$ roles. 
Embedding size was $dSymbols = 10 = dRoles$. 
%This means that the matrix $\mathbf{S}$ in (\ref{eq:TPRforward}) was of size $10 \times 100$ and the matrix $\mathbf{R}$ in (\ref{eq:TPRforward}) was of size $10 \times 20$. 
We used $vec(\mathbf{v}^{(t)})$ as the output of our phrase embedding layer. 
\item The weight of the quantization regularizer in \eqref{Q} was $c_Q = 0.00001$. Results were not highly sensitive to this value.
\item The optimizer used was AdaDelta \cite{adadelta} with 12 epochs.
\end{itemize}

%{\color{red}Each experiment for the \model model took about 13 hours on a single Tesla P100 GPU. We used above setting for our interpretability experiments reported in the next section. Above setting resulted in 2 points performance loss in F1 compared to \cite{BIDAF}. We were able to close this gap by adding gating on the output tensor $\mathbf{v}^{(t)}$ and increasing the number of symbols and roles to 600 and 100 respectively and the embedding size of symbols to 15.  
%Performance results of our single model compared with a single model in \cite{BIDAF} are presented in Table \ref{table:Results}. Please note that the performance reported in Table \ref{table:Results} for \cite{BIDAF} is the result of running the codes published by authors of \cite{BIDAF} with the advised hyperparameters. 
%From this table we observe that our proposed TPR based model outperforms \cite{BIDAF} by 1 point on validation set and slightly underperforms \cite{BIDAF} on test set. 
%This is good enough performance for our proposed \model model to assume that it has learned appropriate representations for questions and paragraphs.
%Focussing on queries, we explore the representations in the learned TPRs in considerable detail. Section \textit{Correcting the Stanford tagger's POS labeling using learned roles} mentions a few selected, highly targeted performance comparisons against the POS tagging provided by the Stanford tagger \cite{StanfParser}.}

Performance results of our model compared to the strong {\sc BiDAF} model proposed in \cite{BIDAF} are presented in Table \ref{table:Results}. We compared the performance of single models. For the {\sc BiDAF} baseline, we ran the code published in \cite{BIDAF} with the advised hyperparameters. Similar to the LSTM used in {\sc BiDAF}, we added a gating mechanism, identical to that of the LSTM cell, to the output tensor $\mb{v}^{(t)}$ in Equation \eqref{eq:TPRforward}. In the TPRN model tested here for performance comparison purposes, we set the number of symbols and roles to 600 and 100 respectively and the embedding size of symbols and roles to 15 and 10. Each experiment for the TPRN model took about 13 hours on a single Tesla P100 GPU. From this table we observe that our proposed TPR based model outperforms \cite{BIDAF} by 1 point on the validation set and slightly underperforms \cite{BIDAF} on the test set. Overall, the proposed TPRN gives results comparable to those of the state-of-the-art {\sc BiDAF} model. Moreover, as we will elaborate in the following sections, our model offers considerable interpretability thanks to the structure built into TPRs.

%%%%%%%%%%%%%%%%Table%%%%%%%%%%%%%%%%%%% 
\begin{table}[h]
\caption{Performance of the proposed \model model compared to {\sc BiDAF} proposed in \cite{BIDAF}}
\label{table:Results}
\begin{center}
\small
\begin{tabular}{  c  c  c  c  c  }
%\hline
Single Model & EM(dev) & F1(dev) & EM(test) & F1(test)\\ \hline
\model & 63.8 & 74.4 & 66.6 & 76.3 \\ %\hline
%{\sc BiDAF} \cite{BIDAF} & 67.11 & 76.8 \\ % The numbers I got by running their code. 
%{\sc BiDAF} \cite{BIDAF} & 62.8 & 73.5 & 67.1  & 76.8  \\ % The numbers reported in their paper.
{\sc BiDAF} & 62.8 & 73.5 & 67.1  & 76.8  \\ % The numbers reported in their paper.
%\hline
\end{tabular}
\end{center}
\end{table}
 %%%%%%%%%%%%%%%%Table%%%%%%%%%%%%%%%%%  
\vspace{-.2in}
\section{Interpretation of learned TPRs}
\label{interpretation}

We separately discuss interpretation of the symbols and the roles learned by \model.

\subsection{Interpreting learned TPR Roles}
\label{subsec:roles}
Here we provide interpretation of the TPR roles $\art$ assigned to the words $w^{(t)}$ of the query input in the forward-directed TPR-RNN of \model. 
(These are denoted $q^{(t)}$ in Fig. \ref{fig:TPR}.)
Just as good learned neural network models in vision typically acquire similar early types of representations of an input image
(e.g., \cite{zeiler2010deconvolutional}), 
it is reasonable to hypothesize that good learned neural network models in language will typically learn low-level input representations that are generally similar to one another. 
Thus we can hope for some generality of the types of interpretation discussed here. 
Convergence on common input representations is expected because these representations capture the regularities among the inputs, useful for many tasks that process such input. 
The kinds of regularities to be captured in linguistic input have been studied for years by linguists, so there is reason to expect convergence between good learned neural network language-input representations and general linguistic concepts. 
The following interpretations provide evidence that such an expectation is merited.

We consider which word tokens $w^{(t)}$ are `assigned to' (or `select') a particular role $k$, meaning that, for an appropriate threshold $\theta_{k}$, $[\mb{\hat{a}_{R}\!}^{(t)}]_{k} > \theta_{k}$  where $\mb{\hat{a}_{R}\!}^{(t)}$ is the $L_{2}$-normalized role-attention vector.
\\

\subsubsection{Grammatical role concepts learned by the model}
\paragraph{A grammatical category---Part of Speech: Determiner $\sim$ Role \#9.}
The network assigns to role \#9 these words:
a significant proportion of the tokens of: \emph{the} (76\%), \emph{an} (52\%), \emph{a} (46\%), \emph{its} (36\%) 
and a few tokens of \emph{of} (8\%) and \emph{Century} (3\%). 
The dominant words assigned to role \#9 (\emph{the, an, a, its}) are all \emph{determiners}.
Although not a determiner, \emph{of} is also an important function word; the 3\% of the tokens of \emph{Century} that activate role \#9 can be put aside.
Quantitatively, $p(w \mr{\ is\ a\ determiner} | w \mr{\ activates\ role\ \#9\ to} > 0.65) = 0.96$.
This interpretation does not assert that \#9 is the \emph{only} role for determiners; e.g., 
$p(w \mr{\ activates\ role\ \#9} | w \in \{a, an, the\}) = 0.70$.

\paragraph{A semantic category: Predicate (verbs and adjectives) $\sim$ Role \#17.}

The words assigned to role \#17 are overwhelmingly predicates, a semantic category corresponding to the syntactic categories of verbs and adjectives [e.g., under semantic interpretation, \emph{J runs $\rightarrow$ runs(J); J is tall $\rightarrow$ tall(J)}] . 
While the English word orders of these two types of predication are often opposite (\emph{the girl runs} vs. \emph{the tall girl}), the model represents them as both filling the same role, which can be interpreted as semantic rather than syntactic.
Quantitatively, $p(w \mr{\ is\ a\ verb\ or\ adjective} | w \mr{\ selects\ role\ \#17}) = 0.82$.
%[activations of role \#17 do not exceed about 0.5, so an activation > 0.25 is taken to mark assignment of role \#17].
Unlike role \#9, which concerns only a small (`closed') class of words, the class of predicates is large (`open'), and role \#17 is assigned to only a rather small fraction of predicate tokens: e.g., $p(w \mr{\ is\ assigned\ to\ role\ \#17} | w \mr{\ is\ a\ verb}) = 0.04$.
	
\paragraph{A grammatical feature: [{\sc plural}] $\sim$ Role \#10.}

To observe the representational difference between the singular and plural roles we need to fix on particular words. 
A case study of \emph{area} vs. \emph{areas} revealed a total separation in their attention to role \#10 (which has a midpoint level of 0.25):
%(because it happens there are numerous tokens of each in the query corpus: 68 and 35, respectively).
%Like role \#17, role \#10's activation never gets larger than about 0.5 in encodings of \emph{area(s)}. 
100\% of tokens of singular \emph{area} have $[\mb{\hat{a}_{R}\!}^{(t)}]_{10} < 0.25$; 100\% of tokens of plural \emph{areas} have  $[\mb{\hat{a}_{R}\!}^{(t)}]_{10} > 0.25$.
This conclusion is corroborated by pronouns, where \emph{he;him} each have mean $[\mb{\hat{a}_{R}\!}^{(t)}]_{10} = 0.1$, while \emph{they;them} have $[\mb{\hat{a}_{R}\!}^{(t)}]_{10} = 0.4;0.6$ (there are very few tokens of \emph{she;her}).

\paragraph{A grammatical phrase-type: \emph{wh-}operator restrictor `phrase' $\sim$ Role \#1.}

Role \#1 is assigned to sequences of words including \emph{how many teams, what kind of buildings, what honorary title}.
We interpret these as approximations to a \emph{wh-restrictor phrase}: a \emph{wh}-word together with a property that must hold of a valid answer---crucial information for question-answering. 
In practice, these `phrases' span from a \emph{wh}-word to approximately the first following content word.
Other examples are:
\emph{what was the American, which logo was, what famous event in history}.

\paragraph{Grammatical functions: Subject/agent vs. object/patient $\sim$ Role \#6.}

A fundamental abstract distinction in syntax/semantics separates subjects/agents from objects/themes. 
In English the distinction is explicitly marked by distinct word forms only on pronouns: \emph{he loves her} vs. \emph{she loves him}. 
In the model, attention to role \#6 is greater for subjects than objects, for both \emph{he} vs. \emph{him} and \emph{they} vs. \emph{them}   
(again, too few tokens of \emph{she;her}).
All but 13 of 124 tokens of \emph{he} and all 77 tokens of \emph{they} allocate high attention to \#6, whereas only 1 of the 27 tokens of \emph{him} and none of the 34 tokens of \emph{them} do (relative to baselines appropriate for the different pairs: 0.52, 0.18). 
\\

\subsubsection{Correcting the Stanford Tagger's POS labeling using learned roles}
\label{sec:correcting}
\paragraph{When \emph{Doctor Who} is not a name: Role \#7.}

The TV character Doctor Who (\emph{DW}) is named many times in the SQuAD query corpus.
Now in \emph{$\ldots$ DW travels $\ldots$} , the phrase \emph{DW} is a proper noun (`NNP'), with unique referent, but 
in \emph{$\ldots$ does the first DW see $\ldots$}, the phrase \emph{DW} must be a common noun (`NN'), with open reference.
In such cases the Stanford tagger misclassifies \emph{Doctor} as an NNP in 9 of 18 occurrences: see Table \ref{CorrectingI}a.
In \emph{$\ldots$ the first DW serial $\ldots$}, \emph{first} modifies \emph{serial} and \emph{DW} is a proper noun. 
The tagger misparses this as an NN in 37 of 167 cases.
Turning to the model, we can interpret it as distinguishing the NN vs. NNP parses of \emph{DW} via role \#7, which it assigns for the NN, but not the NNP, case. 
Of the Stanford tagger's 9 errors on NNs and 37 errors on NNPs, the model misassigns role \#7 only once for each error type (shown in parentheses in Table \ref{CorrectingI}a).
The model makes 7 errors total (Table \ref{CorrectingI}b) while the tagger makes 46.  
Focussing on the specific NN instances of the form \emph{the $n^{th}$ DW}, there are 19 cases:
%(with $n \in \{1, 3, 4, 5, 10, 11\}$)
the tagger was incorrect on 11, and in every such case the model was correct; the tagger was correct in 8 cases and of these the model was also correct on 6.

\begin{table*}[t]
  \small
  \caption{Doctor Who? Correcting the Stanford Tagger (\emph{errors in bold})}
  \label{CorrectingI}
  \centering
  \begin{tabular}{llll|llll}
    \toprule
    {\bf a.} & & \multicolumn{2}{l}{\hspace{.3in}True} & {\bf b.}  & & \multicolumn{2}{l}{\hspace{.3in}True} \\
    &      & NN     & NNP   &   &   & NN     & NNP \\
    \cmidrule{3-4}
    \cmidrule{7-8}
   {\bf tagger} & NN & 9 (5) & \ \ {\bf 37 (1)} & {\bf model} & NN & 13 (5) & \ \ \ \ {\bf 2 (1)}    \\
   (\& model) & NNP & {\bf 9 (1)} & 130 (129) & (\& tagger) & NNP & \, {\bf 5 (1)} & 165 (129)     \\
    \bottomrule
  \end{tabular}
\end{table*}

\paragraph{When \emph{Who} is a name: Role \#1.}

In \emph{Doctor Who travelled}, the word \emph{Who} should not be parsed as a question word (`WP'), but as part of a proper noun (NNP). 
The Stanford tagger makes this error in every one of the 167 occurrences of \emph{Who} within the NNP \emph{Doctor Who}. 
The \model model, however, usually avoids this error. 
Recalling that role \#1 marks the \emph{wh-}restrictor `phrase', we note that in 81\% of these NNP-\emph{Who} cases, the model does not assign role \#1 to \emph{Who} 
(in the remaining cases, it does assign role \#1 as it includes \emph{Who} within its \emph{wh}-restrictor `phrase', generated by a distinct genuine \emph{wh}-word  preceding \emph{Who}).
In all 30 instances of \emph{Who} as a genuine question word in a sentence containing \emph{DW}, the model correctly assigns role \#1 to the question word.
For example, in \emph{Who is the producer of Doctor Who?} [query 7628], the first \emph{Who} correctly selects role \#1 while the second, correctly, does not.
(The model correctly selects role \#1 for non-initial \emph{who} in many cases.)
%(And it is not merely that only sentence-initial \emph{wh-}words select role \#1; in \emph{In addition to a new body, what else changes about the Doctor?} [question 7642], role \#1 marks as the \emph{wh-}restrictor `phrase' the non-sentence-initial \emph{what else changes about}.)

\paragraph{When \emph{to doctor} is not a verb: Role \#17.}

The Stanford tagger parses \emph{Doctor} as a verb in 4 of its 5 occurrences in \emph{$\ldots$ to Doctor Who $\ldots$}.
The model does not make this mistake on any of these 5 cases: it assigns near-zero activity to role \#17,  identified above as the predicate role for verbs and adjectives. 

\subsection{Interpreting learned TPR symbols}
\label{subsec:fillers}

\subsubsection{Meaning of learned symbols: Lexical-semantic coherence of symbol assignments.}
To interpret the lexical-semantic content of the TPR symbols $\mb{s}^{(t)}$ learned by the \model network:
% perform the following steps:
\begin{enumerate}
\item $\mb{s}^{(t)} = \mb{S}\aft \in \R^{10}$ is calculated for all (120,950) word tokens $w^{(t)}$in the validation set.
\item The cosine similarity is computed between $\aft$ and the embedding vector of each symbol.
%, i.e., each of the 100 columns of learned $\mb{S}$.
\item The symbol with maximum (cosine) similarity is assigned to the corresponding token.  
\item For each symbol, all tokens assigned to it are sorted based on their similarity to it; tokens of the same type are removed, and the  top tokens from this list are examined to assess by inspection the semantic coherence of the symbol assignments (see Tables \ref{table:filler27} -- \ref{table:filler6}). 
\end{enumerate}

The results provide significant support for our hypothesis that each symbol corresponds to a particular meaning, assigned to a cloud of semantically-related word tokens. 
For example, symbol 27 
%(all tokens of which are given in Table \ref{table:filler27})
and symbol 6 
%(the top 30 tokens of which are given in Table \ref{table:filler6}) 
can be respectively interpreted as meaning `occupation' and `geopolitical unit'. 
Symbol 11 is assigned to multiple forms of the verb \emph{to be}, e.g., \emph{was} ($85.8\%$ of occurrences of tokens in the validation set), \emph{is}, ($93.2\%$) \emph{being} ($100\%$) and \emph{be} ($98\%$). 
Symbol 29 is selected by 10 of the 12 month names (along with other word types; more details in supplementary materials).
Other symbols with semantically coherent token sets are reported in the supplementary materials. 
Some symbols, however, lack identifiable coherence; an example is presented in Table \ref{table:filler2}. 

%%%%%%%%%%%%%%%%Table%%%%%%%%%%%%%%%%%%% 
\begin{table*}[htbp]
\scriptsize
\parbox[t]{.25\linewidth}{
  \centering
  \caption{\footnotesize Symbol 27}
    \begin{tabular}{ll}    
    \textbf{Token} & \textbf{Similarity} \\
    \hline
    
    printmaker & 0.9587 \\ 
    composer & 0.8992 \\ 
    who   & 0.8726 \\ 
    mathematician & 0.8675 \\ 
    guitarist & 0.8622 \\ 
    musician & 0.8055 \\ 
    Whose & 0.7774 \\ 
    engineer & 0.7753 \\ 
    chemist & 0.7485 \\ 
    how   & 0.7335 \\ 
    strict & 0.7207  \\ 
    \\
    \\
    \end{tabular}%
  \label{table:filler27}%
}
%%%%%%%%%%%%%%%%Table%%%%%%%%%%%%%%%%%%%
\hspace{3pt}
%%%%%%%%%%%%%%%%Table%%%%%%%%%%%%%%%%%%%
\parbox[t]{.25\linewidth}{
  \centering
  \caption{\footnotesize Symbol 2}
    \begin{tabular}{ll}    
    \textbf{Token} & \textbf{Similarity} \\
    \hline
        
    phrase & 0.817  \\
    wrong & 0.8146  \\ 
    mean  & 0.7972  \\
    constitutes & 0.7771  \\
    call  & 0.7621  \\
    happens & 0.752  \\
    the   & 0.7477  \\
    God   & 0.7425  \\
    nickname & 0.7368  \\
    spelled & 0.7162  \\
    name  & 0.712  \\
    happened & 0.6889  \\
    as    & 0.6699  \\
    defines & 0.647  \\
    \end{tabular}%
  \label{table:filler2}%
  }
%%%%%%%%%%%%%%%%Table%%%%%%%%%%%%%%%%%%% 
\hspace{3pt}
%%%%%%%%%%%%%%%%Table%%%%%%%%%%%%%%%%%%% 
  \parbox[t]{.33\linewidth}{
  \centering
  \caption{\footnotesize Symbol 6}
    \begin{tabular}{ll}    
    \textbf{Token} & \textbf{Similarity} \\
    \hline    
    
    abolished & 0.8777  \\
    west  & 0.8734  \\
    nations & 0.8613 \\
    Newcastle & 0.8588 \\
    south & 0.8573 \\
    Melbourne & 0.8558 \\
    Australia & 0.8544 \\
    World & 0.8526 \\
    Belgium & 0.849 \\
    donors & 0.8476 \\
    Asian & 0.8404 \\
    Greece & 0.8402 \\
    Europe & 0.8397 \\
    Thailand & 0.8393 \\
    Constituency & 0.8361 \\
    \end{tabular}%
  \label{table:filler6}% 
    \begin{tabular}{ll}    
    \textbf{Token} & \textbf{Similarity} \\
    \hline    

    annexed & 0.836 \\        
    Brisbane & 0.8359 \\
    European & 0.8341 \\
    Scotland & 0.8321 \\
    Cyprus & 0.8275 \\
    governments & 0.8266 \\
    Commonwealth & 0.8261 \\
    Britain & 0.8243 \\
    flexibility & 0.8227 \\
    territories & 0.8219 \\
    Switzerland & 0.821 \\
    countries & 0.8206 \\
    freedom & 0.819 \\
    Germans & 0.8178 \\
    north & 0.8173 \\
    \end{tabular}%  
  }
\end{table*}%
%%%%%%%%%%%%%%%%Table%%%%%%%%%%%%%%%%%%%

\subsubsection{Polysemy.}
Each token of the same word, e.g., \emph{who}, generates its own symbol/role TPR in \model and if our hypothesis is correct, tokens with different meaning should select different symbols. Indeed we see three general patterns of symbol selection for \emph{who}. 
\emph{Who is the producer of Dr. Who?} illustrates the main-question-word meaning and the proper-name meaning, respectively, in its two uses of \emph{who}. 
Third is the relative pronoun meaning, illustrated by \emph{$\ldots$ the actor who $\ldots$}.
The three symbol-selection patterns associated with these three meanings are shown in Table \ref{table:whos}. 
%(setting aside 4 other symbols which are each assigned to fewer than 8 tokens in total).

\begin{table}[htbp]
\small
  \centering
  \caption{Symbols selected by meaning of \emph{who}}
    \begin{tabular}{lrrrr}
 		Meaning  \hspace{3pt} Symbol ID:	&      			  25 	&  				  52 	&    97 	&    98\\
    \hline
    main question word 								&  						&  \textbf{1062}	&       	& 			\\ 
    relative pronoun 									& 		  \textbf{14}	&   				   4	& 		1	& 		26	\\ 
    proper noun		 									& 							& 					   27 	& 	  16	& 		\textbf{126}	\\ 
	\end{tabular}
  \label{table:whos}
\end{table}
We can interpret the symbols with IDs 25, 52 and 98 as corresponding, respectively, to the meanings of a relative pronoun, a main question word, and a proper noun. 
The tokens with boldface counts are then correct, while the other counts are errors.
Of interest are the further facts that all 18 of the non-sentence-initial main-question-word tokens are correctly identified as such (assigned symbol 52) and that, of the 27 cases of proper-noun-\emph{who}s mislabeled with the main-question symbol 52, half are assigned role \#1, placing them in the \emph{wh}-restrictor `phrase' (whereas only one of the 126 correctly-identified proper-noun-\emph{who}s is).
The Symbol-97-meaning of \emph{who} is at this point unclear.

\subsubsection{Predicting output errors from internal mis-representation.}

In processing the test query \emph{What type/genre of TV show is Doctor Who?} [7632] the model assigns symbol 52 to \emph{Who}, which we have interpreted as an error since symbol 52 is assigned to every one of the 1062 occurrences of \emph{Who} as a query-initial main question-word. 
Although the model strongly tends to give responses of the correct category, here it replies \emph{Time Lord}, an appropriate type of answer to a true \emph{who} question but not to the actual question.
The model makes 4 errors of this type, of the 9 errors total made when assigning symbol 25; this 44\% rate contrasts with the 9\% rate when it correctly assigns the `proper-noun symbol' 98 to \emph{Who}.

Although such error analysis with \model models is in its infancy, it is already beginning to reveal its potential to make it possible, we believe for the first time, to attribute overall output errors of a DNN modeling a language task to identifiable errors of internal representation.
The analysis also shows how the proposed interpretations can be validated by supporting explanations for aspects of the model's behavior.
 
\section{Related work}
\label{related_work}

\paragraph{Architecture.}
In recent years, a number of DNNs have achieved notable success by reintroducing elements of symbolic computation as peripheral modules.
This includes, e.g.: (i) the memory bank, a discrete set of addressed storage registers each holding a neural activation vector 
\cite{henaff2017tracking,sukhbaatar2015end,weston2014memory}; 
and (ii) the sequential program, a discrete sequence of steps, each selected from a discrete set of simple, approximately-discrete primitive operations
\cite{graves2014neural,neelakantan2016neural}.
The discreteness in these peripheral modules is softened by continuous parameters with which they interface with the central controlling DNN;
these parameters modulate (i) the writing and reading operations with which information enters and exits a memory bank (`attention'
 \cite{chorowski2015attention,xu2015show}); 
and (ii) the extent to which inputs are passed to and outputs retrieved from the set of operations constituting a program
\cite{graves2016hybrid}.
The continuity of these parameters is of course crucial to enabling the overall system to be learnable by gradient-based optimization.

The present work constitutes a different approach to reintroducing approximately symbolic representations and rule-based processing into neural network computation over continuous distributed representations.
In computation with TPRs, the symbols and rules are internal to the DNN; there is no separation between a central network controller and peripheral quasi-discrete modules.
Items in memories are distributed representations that are combined by addition/superposition rather than by being slotted into external discrete locations.
Computation over TPRs is massively parallel \cite{THM}.

\paragraph{Interpretation.}
Most methods of interpreting the internal representations of DNNs do so through the input and output representations of DNNs which are by necessity interpretable: these are where the DNN must interface with our description of its problem domain.
An internal neuron may be interpreted by looking at the (interpretable) input patterns that activate it, or the (interpretable) output patterns that it activates
(e.g., \cite{zeiler2014visualizing}).

The method pursued in this paper, by contrast, interprets internal DNN states not via $I \rightarrow O$ behavior but via an abstract theory of the system's problem domain.
In the case of a language processing problem, such theories are provided by theoretical linguistics and traditional, symbolic computational linguistics.
The elements we have interpreted are TPR roles, and TPR fillers, which are distributed activation vectors incorporated into network representations via the summation of their tensor products; we have designed an architecture in which individual neurons localize the presence of such roles and fillers ($\art$ and $\aft$).
Our interpretation rests on the interrelations between activations of the roles and fillers selected to encode words-in-context with the lexical-semantic and grammatical properties attributed to those words-in-context by linguistic theories.

\section{Conclusion}
\label{conclusion}

We introduce a modification of the {\sc BiDAF} architecture for question-answering with the SQuAD dataset. 
This new model, \model, uses Tensor Product Representations in recurrent networks to encode input words.
Through end-to-end learning the model learns how to deploy a set of symbols into a set of structural roles; the symbols and roles have no meaning prior to learning.
We hypothesized that the symbols would acquire lexical meanings and the roles grammatical meanings.
We interpret the learned symbols and roles by observing which of them the trained model selects for encoding individual words in context.
We observe that the words assigned to a given symbol tend to be semantically related, and the words assigned to a given role correlate with abstract notions of grammatical roles from linguistic theory. 
Thus the \model model illustrates how learning to perform a natural language question-answering task can lead a deep learning system to create representations that are interpretable as encoding abstract grammatical concepts without ever being exposed to data labelled with anything like grammatical structure. 
It is widely assumed that it is in such a setting that children learn their first language, so the work lends plausibility to the hypothesis that abstract notions of linguistic theory do in fact describe representations in speakers' minds---representations that are learned in the service of performing tasks that do not explicitly necessitate any such structure.

\bibliography{refs_AAAI2018}

\begin{thebibliography}{}

\bibitem[\protect\citeauthoryear{Abadi \bgroup et al\mbox.\egroup
  }{2015}]{tensorflow2015-whitepaper}
Abadi, M.; Agarwal, A.; Barham, P.; Brevdo, E.; Chen, Z.; Citro, C.; Corrado,
  G.~S.; Davis, A.; Dean, J.; Devin, M.; Ghemawat, S.; Goodfellow, I.; Harp,
  A.; Irving, G.; Isard, M.; Jia, Y.; Jozefowicz, R.; Kaiser, L.; Kudlur, M.;
  Levenberg, J.; Man\'{e}, D.; Monga, R.; Moore, S.; Murray, D.; Olah, C.;
  Schuster, M.; Shlens, J.; Steiner, B.; Sutskever, I.; Talwar, K.; Tucker, P.;
  Vanhoucke, V.; Vasudevan, V.; Vi\'{e}gas, F.; Vinyals, O.; Warden, P.;
  Wattenberg, M.; Wicke, M.; Yu, Y.; and Zheng, X.
\newblock 2015.
\newblock {TensorFlow}: Large-scale machine learning on heterogeneous systems.
\newblock Software available from \url{http://tensorflow.org/}.

\bibitem[\protect\citeauthoryear{Cho and
  Smolensky}{2016}]{cho_bifurcation_2016}
Cho, P.~W., and Smolensky, P.
\newblock 2016.
\newblock Bifurcation analysis of a {{Gradient Symbolic Computation}} model of
  incremental processing.
\newblock In Papafragou, A.; Grodner, D.; Mirman, D.; and Trueswell, J.~C.,
  eds., {\em Proceedings of the 38th {{Annual Conference}} of the {{Cognitive
  Science Society}}}.
\newblock Austin, TX: {Cognitive Science Society}.

\bibitem[\protect\citeauthoryear{Chorowski \bgroup et al\mbox.\egroup
  }{2015}]{chorowski2015attention}
Chorowski, J.~K.; Bahdanau, D.; Serdyuk, D.; Cho, K.; and Bengio, Y.
\newblock 2015.
\newblock Attention-based models for speech recognition.
\newblock In {\em Advances in Neural Information Processing Systems},
  577--585.

\bibitem[\protect\citeauthoryear{Graves \bgroup et al\mbox.\egroup
  }{2016}]{graves2016hybrid}
Graves, A.; Wayne, G.; Reynolds, M.; Harley, T.; Danihelka, I.;
  Grabska-Barwi{\'n}ska, A.; Colmenarejo, S.~G.; Grefenstette, E.; Ramalho, T.;
  Agapiou, J.; et~al.
\newblock 2016.
\newblock Hybrid computing using a neural network with dynamic external memory.
\newblock {\em Nature} 538(7626):471--476.

\bibitem[\protect\citeauthoryear{Graves, Wayne, and
  Danihelka}{2014}]{graves2014neural}
Graves, A.; Wayne, G.; and Danihelka, I.
\newblock 2014.
\newblock Neural turing machines.
\newblock {\em arXiv preprint arXiv:1410.5401}.

\bibitem[\protect\citeauthoryear{Henaff \bgroup et al\mbox.\egroup
  }{2017}]{henaff2017tracking}
Henaff, M.; Weston, J.; Szlam, A.; Bordes, A.; and LeCun, Y.
\newblock 2017.
\newblock Tracking the world state with recurrent entity networks.
\newblock In {\em 6th International Conference for Learning Representations}.

\bibitem[\protect\citeauthoryear{Hochreiter and
  Schmidhuber}{1997}]{hochreiter1997long}
Hochreiter, S., and Schmidhuber, J.
\newblock 1997.
\newblock Long short-term memory.
\newblock {\em Neural computation} 9(8):1735--1780.

\bibitem[\protect\citeauthoryear{Neelakantan, Le, and
  Sutskever}{2016}]{neelakantan2016neural}
Neelakantan, A.; Le, Q.~V.; and Sutskever, I.
\newblock 2016.
\newblock Neural programmer: Inducing latent programs with gradient descent.
\newblock In {\em 5th International Conference for Learning Representations}.

\bibitem[\protect\citeauthoryear{Palangi \bgroup et al\mbox.\egroup
  }{2016}]{hpTASLP}
Palangi, H.; Deng, L.; Shen, Y.; Gao, J.; He, X.; Chen, J.; Song, X.; and Ward,
  R.
\newblock 2016.
\newblock Deep sentence embedding using long short-term memory networks:
  Analysis and application to information retrieval.
\newblock {\em IEEE/ACM Transactions on Audio, Speech, and Language Processing}
  24(4):694--707.

\bibitem[\protect\citeauthoryear{Pennington, Socher, and Manning}{2014}]{GLOVE}
Pennington, J.; Socher, R.; and Manning, C.
\newblock 2014.
\newblock Glove: Global vectors for word representation.
\newblock In {\em Proceedings of the 2014 Conference on Empirical Methods in
  Natural Language Processing (EMNLP)},  1532--1543.

\bibitem[\protect\citeauthoryear{Rajpurkar \bgroup et al\mbox.\egroup
  }{2016}]{squad}
Rajpurkar, P.; Zhang, J.; Lopyrev, K.; and Liang, P.
\newblock 2016.
\newblock {SQuAD}: 100,000+ questions for machine comprehension of text.
\newblock In {\em Proceedings of the 2016 Conference on Empirical Methods in
  Natural Language Processing, {EMNLP}}.
\newblock Available at \url{http://arxiv.org/abs/1606.05250}.

\bibitem[\protect\citeauthoryear{Seo \bgroup et al\mbox.\egroup }{2016}]{BIDAF}
Seo, M.~J.; Kembhavi, A.; Farhadi, A.; and Hajishirzi, H.
\newblock 2016.
\newblock Bidirectional attention flow for machine comprehension.
\newblock In {\em 5th International Conference for Learning Representations}.
\newblock Available at \url{https://arxiv.org/abs/1611.01603}.

\bibitem[\protect\citeauthoryear{Smolensky and Legendre}{2006}]{THM}
Smolensky, P., and Legendre, G.
\newblock 2006.
\newblock {\em The Harmonic Mind: From Neural Computation to
  Optimality-Theoretic Grammar}.
\newblock Cambridge, MA: The MIT Press.

\bibitem[\protect\citeauthoryear{Smolensky, Goldrick, and Mathis}{2014}]{SGM}
Smolensky, P.; Goldrick, M.; and Mathis, D.
\newblock 2014.
\newblock Optimization and quantization in gradient symbol systems: A framework
  for integrating the continuous and the discrete in cognition.
\newblock {\em Cognitive Science} 38:1102--1138.

\bibitem[\protect\citeauthoryear{Smolensky}{1990}]{TPAI}
Smolensky, P.
\newblock 1990.
\newblock Tensor product variable binding and the representation of symbolic
  structures in connectionist networks.
\newblock {\em Artificial Intelligence} 46:159--216.

\bibitem[\protect\citeauthoryear{Smolensky}{2012}]{PhilTr}
Smolensky, P.
\newblock 2012.
\newblock Symbolic functions from neural computation.
\newblock {\em Philosophical Transactions of the Royal Society --- A:
  Mathematical, Physical and Engineering Sciences} 370:3543--3569.

\bibitem[\protect\citeauthoryear{Sukhbaatar \bgroup et al\mbox.\egroup
  }{2015}]{sukhbaatar2015end}
Sukhbaatar, S.; Weston, J.; Fergus, R.; et~al.
\newblock 2015.
\newblock End-to-end memory networks.
\newblock In {\em Advances in neural information processing systems},
  2440--2448.

\bibitem[\protect\citeauthoryear{Toutanova \bgroup et al\mbox.\egroup
  }{2003}]{StanfParser}
Toutanova, K.; Klein, D.; Manning, C.; and Singer, Y.
\newblock 2003.
\newblock Feature-rich part-of-speech tagging with a cyclic dependency network.
\newblock In {\em Proceedings of HLT-NAACL 2003},  252--259.
\newblock Available at \url{https://nlp.stanford.edu/software/tagger.shtml}.

\bibitem[\protect\citeauthoryear{Weston, Chopra, and
  Bordes}{2014}]{weston2014memory}
Weston, J.; Chopra, S.; and Bordes, A.
\newblock 2014.
\newblock Memory networks.
\newblock {\em arXiv preprint arXiv:1410.3916}.

\bibitem[\protect\citeauthoryear{Xu \bgroup et al\mbox.\egroup
  }{2015}]{xu2015show}
Xu, K.; Ba, J.; Kiros, R.; Cho, K.; Courville, A.; Salakhudinov, R.; Zemel, R.;
  and Bengio, Y.
\newblock 2015.
\newblock Show, attend and tell: Neural image caption generation with visual
  attention.
\newblock In {\em International Conference on Machine Learning},  2048--2057.

\bibitem[\protect\citeauthoryear{Zeiler and
  Fergus}{2014}]{zeiler2014visualizing}
Zeiler, M.~D., and Fergus, R.
\newblock 2014.
\newblock Visualizing and understanding convolutional networks.
\newblock In {\em European conference on computer vision},  818--833.
\newblock Springer.

\bibitem[\protect\citeauthoryear{Zeiler \bgroup et al\mbox.\egroup
  }{2010}]{zeiler2010deconvolutional}
Zeiler, M.~D.; Krishnan, D.; Taylor, G.~W.; and Fergus, R.
\newblock 2010.
\newblock Deconvolutional networks.
\newblock In {\em Computer Vision and Pattern Recognition (CVPR), 2010 IEEE
  Conference on},  2528--2535.
\newblock IEEE.

\bibitem[\protect\citeauthoryear{Zeiler}{2012}]{adadelta}
Zeiler, M.~D.
\newblock 2012.
\newblock {ADADELTA:} an adaptive learning rate method.
\newblock Available at \url{http://arxiv.org/abs/1212.5701}.

\end{thebibliography}
\bibliographystyle{aaai}

\clearpage
\newpage

\section*{\huge Supplementary Material}

\hspace{5cm}

\section{More examples for lexical-semantic coherence of symbol assignments}
\label{sec:vis}

In this section we present more examples that supports the lexical-semantic coherence of the words assigned to symbols described in the section \textit{lexical-semantic coherence of symbol assignments} of the paper. 

\subsection{Symbol 29: ``\textit{month of the year}'' symbol}
Symbol 29 is attracted to the months of the year. By counting the total number of times a month of the year token has appeared in the validation set, and then counting how many of them are attracted to symbol 29, we will get results in Table \ref{table:filler29-a}. Top 30 tokens attracted to this filler are presented in Table \ref{table:filler29-b} (sorted based on cosine similarity, duplicate tokens removed). As observed many of them show a year token.  
%%%%%%%%%%%%%%%%Table%%%%%%%%%%%%%%%%%%% 
\begin{table*}[htbp]

\parbox[t]{.25\linewidth}{
  \centering
  \caption{\footnotesize \textbf{Percentage of occurrences of each month of year token in the whole validation set attracted to symbol 29}}
    \begin{tabular}{lr}    
    \textbf{Token} & \textbf{Attracted tokens} \\
    \hline
    
    January  & 29\% \\ 
    Feb & 100\% \\ 
    February    & 50\% \\ 
    March  & 100\% \\ 
    April  & 71\% \\ 
    May  & 0\% \\ 
    June  & 100\% \\ 
    July  & 44\% \\ 
    August  & 60\% \\ 
    September   & 0\% \\ 
    October  & 71\%  \\
    November & 60\% \\
    December & 25\% \\ 
    \\
    \\
    \end{tabular}%
  \label{table:filler29-a}%
}
%%%%%%%%%%%%%%%%Table%%%%%%%%%%%%%%%%%%%
%\hspace{3pt}
\hfill
%%%%%%%%%%%%%%%%Table%%%%%%%%%%%%%%%%%%%
\parbox[t]{.6\linewidth}{
  \centering
  \caption{\footnotesize \textbf{Symbol 29, top 30 tokens}}
    \begin{tabular}{lr}    
    \textbf{Token} & \textbf{Similarity} \\
    \hline
        
    1862 & 0.8785 \\
    1846 & 0.8711 \\
    1850& 0.8615 \\
    kilometre & 0.8528 \\
    1857 & 0.8506 \\
    kilometer & 0.8492 \\
    1785 & 0.8454 \\
    mile  & 0.8397 \\
    mainline & 0.8346 \\
    slavery & 0.8345 \\
    1942 & 0.8331 \\
    1944 & 0.8303 \\
    1898 & 0.8263 \\
    1914 & 0.8261 \\
    1808 & 0.815 \\
    \end{tabular}    
    \label{table:filler29-b}
    \begin{tabular}{lr}    
    \textbf{Token} & \textbf{Similarity} \\
    \hline
        
    1915 & 0.8144 \\
    War   & 0.8113 \\
    1775 & 0.8067 \\
    1962 & 0.8003 \\
    March & 0.7993 \\
    1755 & 0.7992 \\
    1953 & 0.7934 \\
    sin   & 0.7917 \\
    1965 & 0.7866 \\
    Feb   & 0.7848 \\
    Orleans & 0.7847 \\
    1964 & 0.7842 \\
    18 & 0.7839 \\
    Rhine & 0.7836 \\
    law   & 0.771 \\
    \end{tabular}%
  }
\end{table*}
%%%%%%%%%%%%%%%%Table%%%%%%%%%%%%%%%%%%%

\subsection{Symbol 26: ``\textit{what}'' symbol}
100\% of ``\textit{what}'' and ``\textit{What}'' tokens are attracted to symbol 26. 

\subsection{Symbol 20: ``\textit{directional / causal}'' symbol}
75.8\% of ``\textit{to}'' tokens and 81.7\% of ``\textit{from}'' tokens are attracted to symbol 20.

\subsection{Symbol 55: ``\textit{finance / property}'' symbol}
Most of the tokens attracted to symbol 55 are finance or property related tokens. Table \ref{table:filler55} shows the full list of tokens attracted to symbol 55.

%%%%%%%%%%%%%%%%Table%%%%%%%%%%%%%%%%%%%
\begin{table}[htbp]
  \centering
  \caption{\footnotesize \textbf{Symbol 55, all tokens}}
    \begin{tabular}{lr}
        \textbf{Token} & \textbf{Similarity} \\
    \hline
    
    price & 0.8956 \\
    city  & 0.8674 \\
    town  & 0.8525 \\
    value & 0.8293 \\
    capital & 0.8278 \\
    country & 0.8231 \\
    district & 0.7878 \\
    prices & 0.7636 \\
    nation & 0.7623 \\
    producer & 0.7554 \\
    rate  & 0.7522 \\
    \end{tabular}%
    \begin{tabular}{lr}
        \textbf{Token} & \textbf{Similarity} \\
    \hline
    village & 0.7511 \\
    property & 0.733 \\
    court & 0.7196 \\
    's    & 0.7167 \\
    artist & 0.7164 \\
    product & 0.7054 \\
    per   & 0.7036 \\
    detective & 0.6887 \\
    borough & 0.6869 \\
    store & 0.6705 \\
    income & 0.6632 \\
    \end{tabular}%
  \label{table:filler55}%
\end{table}%
%%%%%%%%%%%%%%%%Table%%%%%%%%%%%%%%%%%%%

\subsection{Symbol 43: ``\textit{how}'' symbol}
100\% of ``\textit{How}'' tokens and 62.6\% of ``\textit{how}'' tokens are attracted to symbol 43.

\subsection{Symbol 22: ``\textit{law}'' symbol}
Most of the tokens attracted to symbol 22 are law related tokens. Table \ref{table:filler22} shows the full list of tokens attracted to symbol 22.

%%%%%%%%%%%%%%%%Table%%%%%%%%%%%%%%%%%%%
\begin{table}[htbp]
  \centering
  \caption{\footnotesize \textbf{Symbol 22, all tokens}}
    \begin{tabular}{lr}
            \textbf{Token} & \textbf{Similarity} \\
    \hline
    
    embargo & 0.8496 \\
    constitution & 0.8329 \\
    article & 0.8313 \\
    amendment & 0.8263 \\
    convicted & 0.8045 \\
    Draft & 0.7623 \\
    Nobel & 0.7568 \\
    Pro   & 0.7566 \\
    Protestant & 0.7552 \\
    \end{tabular}
    \begin{tabular}{lr}
            \textbf{Token} & \textbf{Similarity} \\
    \hline
    
    renounced & 0.7459 \\
    senate & 0.7454 \\
    opposes & 0.7394 \\
    Law   & 0.734 \\
    campaign & 0.7291 \\
    signed & 0.7236 \\
    bridges & 0.6841 \\
    supporting & 0.6613 \\
      &  \\
    \end{tabular}%
  \label{table:filler22}%
\end{table}%
%%%%%%%%%%%%%%%%Table%%%%%%%%%%%%%%%%%%%

\subsection{Symbol 44: ``\textit{territory}'' symbol}
Most of the tokens attracted to symbol 44 are territory related tokens. Table \ref{table:filler44} shows the full list of tokens attracted to symbol 44.
%%%%%%%%%%%%%%%%Table%%%%%%%%%%%%%%%%%%%
\begin{table}[htbp]
  \centering
  \caption{\footnotesize \textbf{Symbol 44, all tokens}}
    \begin{tabular}{lr}
   \textbf{Token} & \textbf{Similarity} \\
    \hline
    
    continents & 0.7916 \\
    regions & 0.7744 \\
    channels & 0.7539 \\
    Bank  & 0.7425 \\
    Regency & 0.7337 \\
    areas & 0.7264 \\
    countries & 0.7258 \\
    \end{tabular}
    \begin{tabular}{lr}
   \textbf{Token} & \textbf{Similarity} \\
    \hline
    
    divisions & 0.7173 \\
    territory & 0.7114 \\
    kingdom & 0.7099 \\
    Duchy & 0.7013 \\
    other & 0.6901 \\
    region & 0.6588 \\
      &  \\
    \end{tabular}%
  \label{table:filler44}%
\end{table}%
%%%%%%%%%%%%%%%%Table%%%%%%%%%%%%%%%%%%%

\subsection{Symbol 61: ``\textit{year}'' symbol}
Most of the tokens attracted to symbol 61 are different year number tokens. Table \ref{table:filler61} shows the full list of tokens attracted to symbol 61.
%%%%%%%%%%%%%%%%Table%%%%%%%%%%%%%%%%%%%
\begin{table}[htbp]
  \centering
  \caption{\footnotesize \textbf{Symbol 61, all tokens}}
    \begin{tabular}{lr}
      \textbf{Token} & \textbf{Similarity} \\
    \hline
    
    1897 & 0.858 \\
    Maroons & 0.839 \\
    1656 & 0.8277 \\
    1954 & 0.8228 \\
    1921 & 0.8195 \\
    1784 & 0.8145 \\
    1901 & 0.8134 \\
    1890 & 0.8104 \\
    1896 & 0.8077 \\
    1891 & 0.8002 \\
    1643 & 0.7892 \\
    1781 & 0.7859 \\
    1756 & 0.7814 \\
    \end{tabular}
     \begin{tabular}{lr}
      \textbf{Token} & \textbf{Similarity} \\
    \hline
    
    17th  & 0.7798 \\
    1953 & 0.7793 \\
    1961 & 0.7707 \\
    1884 & 0.7667 \\
    Irish & 0.766 \\
    1945 & 0.763 \\
    1952 & 0.7604 \\
    1763 & 0.7542 \\
    ISIL  & 0.7509 \\
    1932 & 0.7439 \\
    1951 & 0.7329 \\
    1754 & 0.7316 \\
    British & 0.6596 \\
    \end{tabular}%
  \label{table:filler61}%
\end{table}%
%%%%%%%%%%%%%%%%Table%%%%%%%%%%%%%%%%%%%

\subsection{Symbol 36: ``\textit{a / an / about}'' symbol}
70.8\% of ``\textit{a}'' tokens, 69.7\% of ``\textit{an}'' tokens and 87\% of ``\textit{about}'' tokens are attracted to symbol 36.

\subsection{Symbol 30: ``\textit{?}'' symbol}
74\% of ``\textit{?}'' tokens are attracted to symbol 30.

\end{document}